# CNN-based repetitive self-revised learning for photos' aesthetics imbalanced classification

Ying Dai
Faculty of software and information science
Iwate prefectural university
Takizawa, Japan
dai@iwate-pu.ac.jp

*Abstract*— Aesthetic assessment is subjective, and the distribution of the aesthetic levels is imbalanced. In order to realize the auto-assessment of photo aesthetics, we focus on using repetitive self-revised learning (RSRL) to train the CNN-based aesthetics classification network by imbalanced data set. As RSRL, the network is trained repetitively by dropping out the low likelihood photo samples at the middle levels of aesthetics from the training data set based on the previously trained network. Further, the retained two networks are used in extracting highlight regions of the photos related with the aesthetic assessment. Experimental results show that the CNN-based repetitive self-revised learning is effective for improving the performances of the imbalanced classification.

*Keywords*— *photo aesthetic assessment, repetitive self-revised deep learning, dropping out sample, imbalanced classification, transfer learning, highlight*

I. INTRODUCTION

In response to the growth of digital camera, more and more pictures are taken to upload the social media. Many people hope to improve their aesthetic levels by taking beautiful photographs. So, auto-assessment of photo aesthetics is challenging. Researches have been investigating methods for providing automated aesthetical evaluation and classification of photographs. Aesthetic assessment is subjective. One of the main difficulties in addressing this challenge is in developing formal models of human aesthetic preference [1]. In this paper, authors stated that such models would allow computer systems to predict the aesthetic taste of a human being or adapt to the aesthetic tendencies of a human group. In [2], recent computer vision techniques used in the assessment of image aesthetic quality were reviewed. In [3], a set of features derived from both low- and high-level analysis of photo layout were exploited to perform the aesthetic quality evaluation by a Support Vector Machine (SVM) classifier. In [4], authors designed a set of compact rule-based features based on photographic rules and aesthetic attributes, and used Deep Convolutional Neural Network (DCNN) descriptor to implicitly describe the photo quality. These approaches focused on extracting the handcrafted image features. However, the effectiveness is limited that extracting the features is based on the researchers' understanding on the aesthetic rules. In [5], a scene convolutional layer was designed to learn specific aesthetic features for various scenes by deep learning model. In [6], a novel photograph aesthetic classifier with a deep and wide CNN for fine-granularity aesthetical quality prediction was introduced. In [7], the percentage distributions for orientation, curvature, color and global symmetry were extracted and fed to a deep neural network under the form of only 114 inputs. However, the issue whether the handcrafted features are generic for the photo aesthetic assessment is not involved. Moreover, all of the above approaches were not involved in the issue that the aesthetic rating is ambiguous and is different from person to person, which caused a highly imbalanced distribution of aesthetic ratings. Toward to tackling these issues, authors in [8] showed how to learn deep features for imbalanced data classification. Using the learned features, the classification was simply achieved by a fast cluster-wise kNN search followed by a local large margin decision. In [9], authors proposed an end-to-end CNN model which simultaneously implements aesthetic classification and understanding. A sample-specific classification method that re-weights samples' importance is implemented, and what is learned in the deep model was investigated. Ambiguous samples are given lower weights while clear samples are weighted high. However, the method to give the weight of every sample was not explicit, and the improvement for the imbalanced data classification was not salient from the experiment results.

Motivated by the above research, we collected about 3100 photos scored aesthetically by a professional photographer. These photos were taken by the students of the photographer's class. The scores are in the range of [2, 9]. The photos with score 2 or less are aesthetically poor; those with scores from 3 to 5 are fair; those with scores 6 and 7 are good; those with scores larger than 7 are excellent. The data set indeed exhibited a highly non-uniform distribution against scores as illustrated in Fig.1. Most images concentrate on the scores of 3 to 5 (about 87%). The classification model could be overwhelmed by those general samples if the parameters are learned by treating all samples equally, and the more discriminative samples couldn't decide how the model is trained.

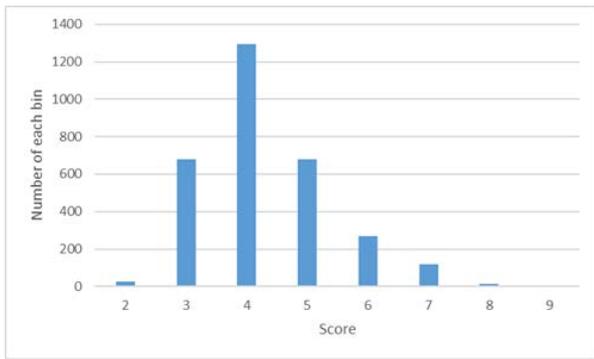

Fig. 1 Score distribution of photo data set

Accordingly, in order to solve the data imbalance issue in aesthetic assessment, in this paper, we focus on using repetitive self-revised learning to train the CNN-based aesthetics classification model by the imbalanced data set. As the repetitive self-revised learning, the network is trained repetitively by dropping out the low likelihood photo samples scored in the range of [3, 5] from the training data set based on the previously trained model. Further, the aesthetic highlight region of photo images are extracted by subtracting two specific feature maps of first convolutional layers of the two retrained models, to analyze the correlations of the highlight regions with the aesthetic assessments. Experimental results show that the CNN-based repetitive self-revised learning is effective for solving the issues of imbalanced classification.

## II. RELATED WORKS

The photos' aesthetic level assessment exhibits highly-skewed score distribution as shown in Fig. 1. As described in [8], for such class-imbalanced data, the minority class often contains very few instances with high degree of visual variability. The scarcity and high variability make the genuine neighborhood of these instances easy to be invaded by other imposter nearest neighbors. In [10], a comprehensive literature survey to tackle the class data imbalance problem was reviewed. Generally, there are two groups of solutions: data re-sampling and cost-sensitive learning. A well-known issue with over-sampling is its tendency to overfitting. Therefore, under-sampling is often preferred, although potentially valuable information may be removed. In [8], a data structure-aware deep learning approach with build-in margins for imbalanced classification was proposed. However, how to utilize these approaches to solve the class data imbalance problem regarding aesthetic assessment is unknown.

## III. REPETITIVE SELF-REVISED LEARNING

In this paper, we propose a CNN-based repetitive self-revised learning (RSRL) method for photos' aesthetics classification by repetitively dropping out the low likelihood samples of majority classes with regards of scores, so as to ameliorate the invasion to the minority classes, and the loss of the samples with discriminative features in the majority classes. The idea behind is the assumption that the sample with low likelihood in the majority classes is what is ambiguously assessed. These samples are easy to invade the genuine neighborhood of samples in the minority classes. The system diagram is shown in Fig. 2.

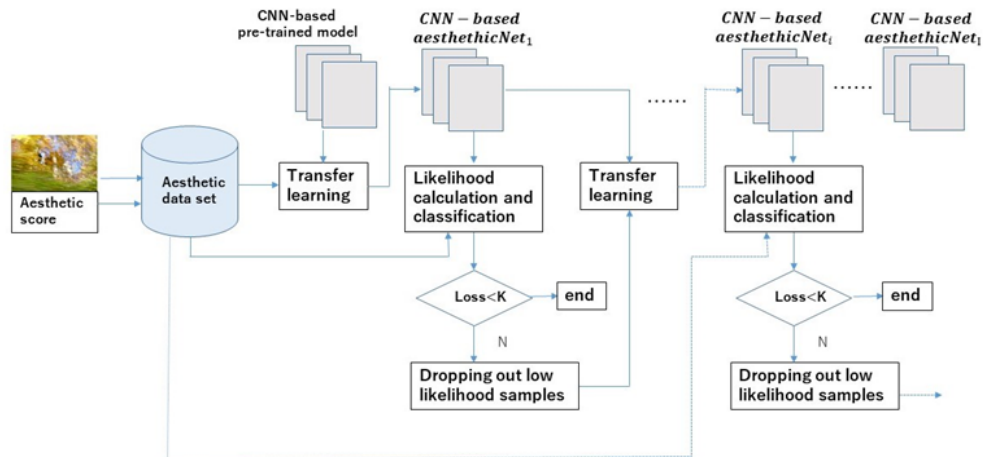

Fig. 2 Diagram of training photo Aesthetic assessment model

The training data set is the score-labeled photo data set. The aesthetic scores in the range of [1, N] are given by a pro-photographer. The scores' distribution of samples in the data set is as shown in Fig. 1. The samples are almost concentrated in the fair classes with score 3 to 5. Score $n$ is handled as one class which is denoted $s_n$, while the number of classes are $N$.

The photos' aesthetics imbalanced classification is tackled by the CNN-based classification network. Network training begins from the pre-trained network such as alexNet by transfer learning. The last three layers of the alexNet are tuned for the score classes. By replacing the last three layers of the alexNet, the network to classify photos instead are fine-tuned by feeding the training data set. The generated network is called $aestheticNet_1$. The network architecture is as the following.

*1-22 layers: alexNet layers' Transferring*

*23 layers 'fc': full connected layer, N nodes, each corresponding to one class*

*24 layer 'softmax': softmax layer, N nodes*

*25 layer 'classoutput': classification output*

Then, the nodes of 'fc' layer of all samples in the training data set are activated to get the Sigmoidal fuzzy membership values, denoted $fc_{s_n}$. The value of $fc_{s_n}$ of the sample could be treated as its likelihood belong to $s_n$. Based on $fc_{s_n}$, the samples in the majority classes, which have low likelihood to $s_n$, are dropped out from the training data set. Such dropping out the low likelihood samples is considered to be a process of self-revision of the training data set. The idea behind is that the samples labeled with the scores of the majority classes while having the low likelihood to them may become the imposter nearest neighbors of the samples in the minority classes to invade the genuine neighborhood of the minority classes. Accordingly, the conditions of dropping-out the low likelihood samples of the majority classes are expressed by the following equations.

*Droping out* a sample $i_{s_{max1}}, if\ its\ fc_{s_{max1}} < K1$ (1)

*Droping out a* sample $i_{s_{max2}}, if\ its\ fc_{s_{max2}} < K2$ (2)

*Droping out a* sample $i_{s_{max3}}, if\ its\ fc_{s_{max3}} < K3$ (3)

Where, the class having most samples is denoted $s_{max1}$, the next is $s_{max2}$, the third is as $s_{max3}$. The sample which is labeled with $s_{max1}$, denoted $i_{s_{max1}}$, is removed from the training data set if the corresponding $fc_{s_{max1}}$ is less than *K1*. And so on, the sample which is labeled with $s_{max2}$ or $s_{max3}$, denoted $i_{s_{max2}}$ or $i_{s_{max3}}$, is dropped out from the training data set if the corresponding $fc_{s_{max2}}$ is less than *K2*, or $fc_{s_{max3}}$ is less than *K3*.

Then, based on the previously trained network $aestheticNet_1$, the network is retrained by transfer learning using the self-revised training data set by dropping out the low likelihood samples. The accordingly generated network is called $aestheticNet_2$. Based on $aestheticNet_2$, the likelihoods of all samples of the original training data set belonging to each class is calculated. The samples are dropped out if they apply to the above removing conditions (1), (2) and (3). It is noticed that the samples that are dropped out based on the previously trained network could be remained in the currently self-revised training data set based on the current network, so as to void in removing the samples with discriminative features for classification. We call such learning process as self-revised learning.

Repetitively, based on the latest retrained network $aestheticNet_{i-1}$, the network $aestheticNet_i$ is retrained by transfer learning with the latest training data set dropping out the low likelihood samples of major classes based on $aestheticNet_{i-1}$. This learning procedure is continued until the total F-measure of the classes regarding the test data set reaches the optimal value. The corresponding generated network is called $aestheticNet_I$, which is utilized as the photos' aesthetics imbalanced classification model.

IV. EXTRACTING AESTHETICS HIGHLIGHT REGION

The retrained CNN-based networks are used to extract aesthetic highlight regions of photos, so as to analyze how the photos are assessed by the pro-photographer to investigate the composition features of good photos, and the correlation of the salient objects with backgrounds. The diagram of the proposed method is shown in Fig. 3.

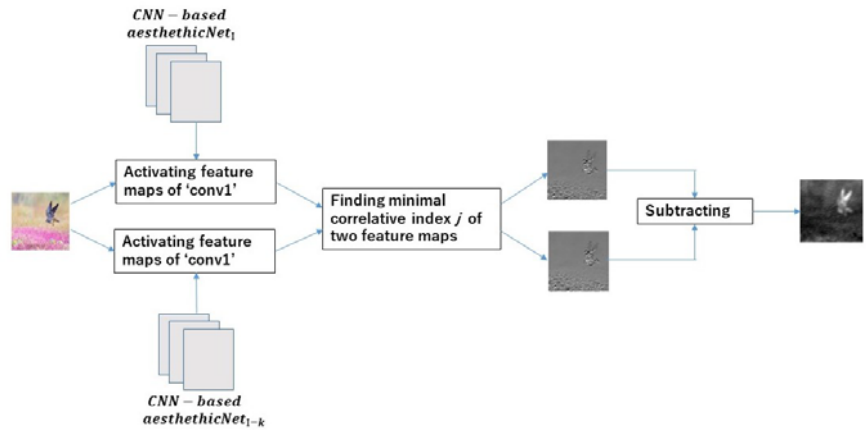

Fig. 3 Diagram of extracting highlight region

For a photo image, feature maps of 'conv1' layer of two networks are activated. These two networks are the latest retrained network $aestheticNet_I$ and the previously retrained network $aestheticNet_{I-k}$. The feature maps are denoted by $conv1_I^j$ and $conv1_{I-k}^j$, respectively. *I-k* indicates the *(I-k)th* retrained network, and *j* does the *j*th feature map. It is assumed that the feature difference map of the two corresponding feature maps, which are of the minimal correlation, could reflect the aesthetics highlight region. The idea behind is that the training data set is aesthetically labeled by the pro-photographer who often focuses on the aesthetics highlight regions which embody photo's aesthetic level based on the essential principles, such as whether the object in the photo is distinctive, and whether the composition is concise. So, the activated values of the highlight regions should change more greatly if the network is retrained by removing the low likelihood samples. Therefore, the feature difference map could be used to extract such aesthetic highlight region.

The index $J$ of the feature maps which are of the minimal correlation is identified by the following equation.

$$J = \arg min_j (corr[conv1_I^j, conv1_{I-k}^j]) \quad (4)$$

Where, *corr* indicates the correlation. The feature difference map of the two feature maps with index $J$ could be calculated by subtracting $conv1_I^J$ and $conv1_{I-k}^J$, as the equation below.

$$diff_{I,I-k} = conv1_I^J - conv1_{I-k}^J \quad (5)$$

Fig.4 shows the feature difference maps of two photos. The left column are original images; the middle are feature maps regarding $conv1_{10}^J$ and $conv1_9^J$, respectively; the right are the feature difference maps of them, denoted $diff_{10,9}$.

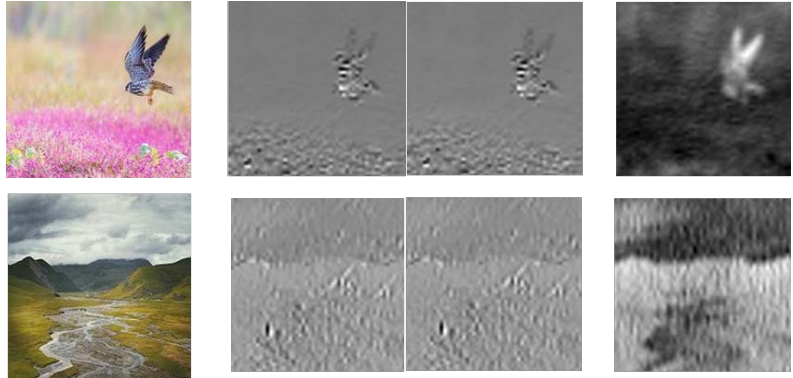

Fig. 4 Calculating feature difference map

For the upper example, the salient object bird as a highlight is emerged explicitly in the feature difference map; for the lower example, the mountain area is emerged in the feature difference map although the highlight of this sample is not obvious.

V. EXPERIMENTS AND ANALYSIS

A. *Data Set*

Although the AVA data set [11] is the largest publicly available aesthetics dataset providing over 250,000 images in total, each image in which was aesthetically assessed by about 200 people with the rating score ranging from 1 to 10, all of the images were finally labeled with the mean score that lost the individual`s aesthetic sense although the aesthetic tendencies of a human group could be reflected. However, embodying the aesthetic taste of a human being is important in training our aesthetics imbalanced classification network. Therefore, we conduct our data set which contains 3100 photos assessed aesthetically by a professional photographer, which is called xiheAA. These photos were taken by the students of the photographer's class. The scores range from 2 to 9. The distribution of the scores is shown in Fig. 1. The class having most samples is score 4; the next is score 3; the third is score 5.

For the 5-fold cross validation, four of fifth samples are selected randomly as the training dataset, and the rest samples are used as the test data set. That is, the size of training data set is 2480 samples, and the size of test data set is 620 samples.

B. *Aesthetics imbalanced classification*

In this section, we evaluate how the RSRL improve the photo aesthetics imbalanced classification performance. The approach proposed in section 3 is implemented on Matlab. The function *trainNetwork* is used repetitively to fine-tune the weights of the CNN-based pre-trained network by inputting the self-revised training data set to obtain the novel classification network. The epoch is set as 5. The function *activation* is used to activate the nodes values of layer 'fc' to obtain the likelihoods of samples to each score class. The function *classify* is used to assign the test samples to the corresponding score classes using the retrained classification networks.

Fig.4 shows the change of sizes of the training data set caused by RSRL. The conditions dropping out the low likelihood samples are based on the expressions (1), (2), and (3). There, the values of K1, K2, and K3 are firstly set as 0.9, respectively. After the 2th round of RSRLs, the values of K1, K2, and K3 are adjusted as 0.95, appropriately.

The initial size of the self-revised training data set is about 1300 samples for the first round self-revised learning. About 1/2 samples in the original training data set are removed, including the samples with discriminative features. Then for the second round RSRL, about 900 samples are remained in the training data set. After that, the sizes increase gradually with RSRL. That means, the some samples with discriminative features not invading the genuine neighborhood of the minority classes come back in the self-revised training data set by RSRL. When the number of times of RSRLs are larger than 20, the sizes of the training data set are little changed, being stale at about 1800 samples. the sharing rate of the major classes with score 3 to score 5 is 87% for the original training data set, while the sharing rate becomes 83% for the 20th round training data set.

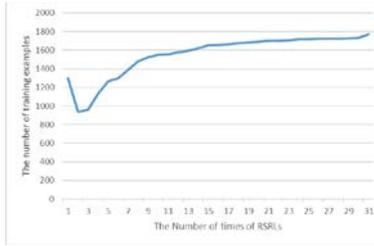

Fig.4 Change of sizes of training data set with RSRL

The classification performances of the generated networks are evaluated by the precision, recall and F-measure. The F-measure is calculated by the following equation (6).

$$F-\text{measure} = \frac{2*precision*recall}{precision+recall} \quad (6)$$

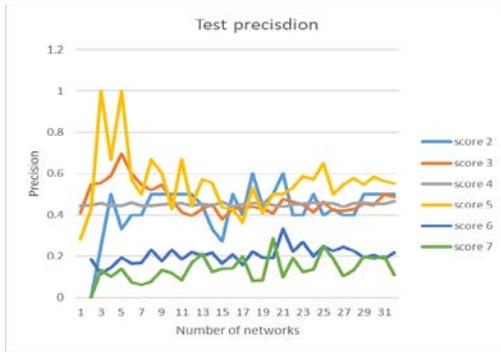

Fig. 5 Test precision

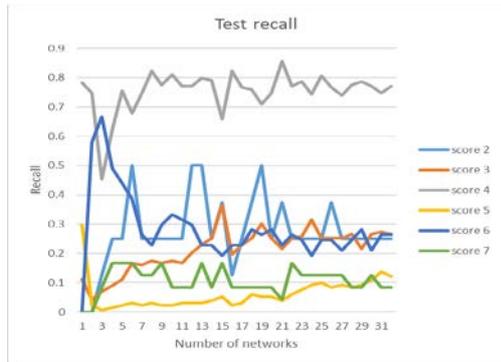

Fig. 6 Test recall

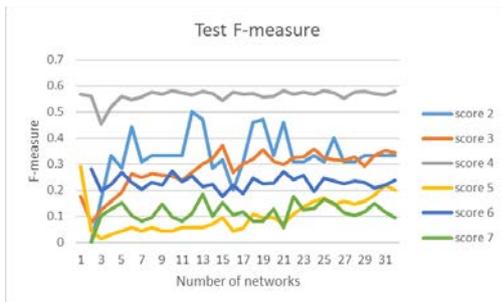

Fig.7 Test F-measure

The precisions, recalls and F-measures of thirty classification networks retrained iteratively by RSRL for each class of the test data set are show in Fig. 5, fig. 6 and fig. 7. Number 1 corresponds to the initial network, Number 2 does the second retrained network, and so on. There are not samples of score 9 in the test data set, because only 2 samples were collated in the xiheAA data set. Also, there are only 2 samples of score 8 in the test data set. So, the results with regards to the class score 8 and score 9 are not shown in the figures.

It is observed from the results of Fig.5, fig.6 and fig.7 that the precision, recall and F-measure of each class are improved gradually with RSRL. For most of the classes, the changes of performance become stable after the 20th network. For the majority class score 4 which occupies about the 1/2 of the training dada set, the classification performances are stable, although there are a little fluctuation in the initial steps of RSRL for the recall and F-measure. For the minority class of score 2, the performances also become stable, although there are a little severe fluctuations with RSRL. Fig.8 shows the change of total F-measure values regarding the classes. It is observed that the total F-measure is optimal for the 24th network. Accordingly, this retrained network could be considered the optimal network for the photos' aesthetics imbalanced classification.

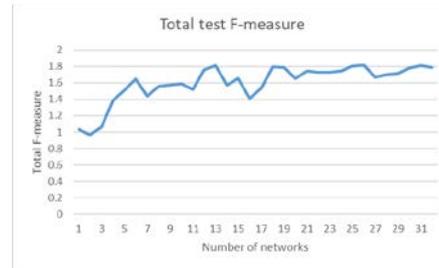

Fig.8 Change of total F-measures with RSRL

Moreover, it is observed that there are no samples in the test data set which are assigned to the minority classes of score 2, score 6 and score7 for the initial network, although there are the samples labelled to the minority classes. However, with RSRL, the retrained networks begin to assign some of these samples to the minority classes. For these minority classes, the average precision, average recall and average F-measure of the 24th retrained network increase to 0.3, 0.2 and 0.24, respectively. The improvement is great, although the values are not so high.

For the majority classes of score 3, score 4 and score 5, the average precision, average recall and average F-measure of the initial network are 0.38, 0.40, and 0.34, and those of the 24th network are 0.52, 0.39 and 0.36, respectively. The improvement of precision is obvious.

As the whole, from the above experimental results, we can see that the performance of the imbalanced classification could be improved by the CNN-based RSRL.

Further, the accuracy is calculated by the following equation (7).

$$\text{accuracy} = \frac{\sum \#TP_i}{\#all} \quad (7)$$

Where, # indicates the number, $\#TP_i$ indicates the number of photos assigned to class $i$ correctly, and $\#all$ indicates the number of photos in the data set.

For the 24th retrained network, the accuracy of the test data set is 0.45. However, the accuracy of the training data set is 0.99. We can see that the overfitting becomes severe with RSRL. We think that increasing the samples of the minority classes appropriately could improve overfitting on some degree.

Some examples assigned to the classes of score 3 and score 7 are shown in Fig. 9. The upper images are assigned to the score 3, and the lower images are assigned to score 7.

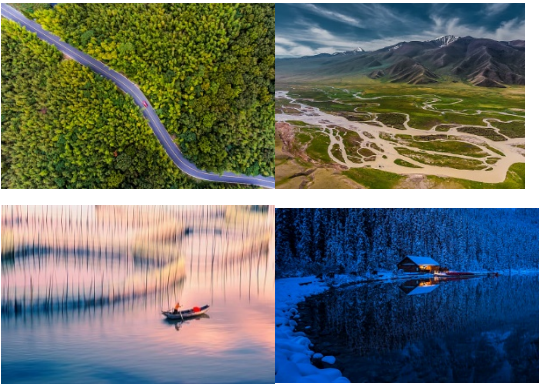

Fig. 9 Examples assigned to score 3 and score 7

It is obviously that the visual aesthetic quality of the lower images is better than one of the upper images, and seems to meet the common techniques for composing a good photo. So, the classification results are reliable.

Accordingly, we can say that RSRL for training the CNN-based network improves the performances of the photos' aesthetics classification. The experimental results verify that the issues of imbalanced classification could be solved by RSRL. However, with RSRL, the severe overfitting is occurred. It is necessary to research more in the future.

*C. Aesthetic analysis of highlight region*

In section 4, extracting aesthetics highlight region from the photo image by using the repetitively trained networks was proposed. In this section, we illustrate how this method reveals the correlation of the extracted region with the aesthetic assessment.

Fig.10 shows the extracted highlight regions of some photos labeled with score 7. The left is the set of some original images; the right is the set of the corresponding feature difference maps calculated by 9th network and the 10th network.

It is observed that feature difference maps are very clear. The salient objects are distinctive, and look pretty. It seems that the photos' highlight regions extracted by using the retrained aesthetics classification networks reveal the photo's aesthetic qualities. By analyzing the compositions of the extracted elements with the aesthetic scores, it is possible to learn how to arrange the elements in the photo to make up a good photo.

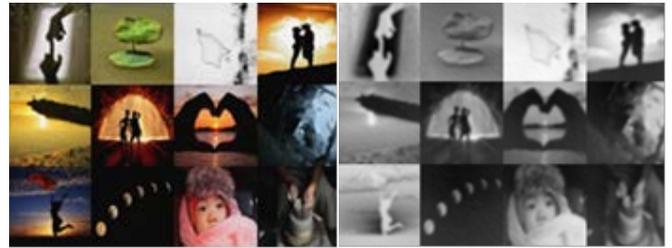

Fig. 10 Highlight regions of some photos labeled with score 7

VI. CONCLUSION

In this paper, we proposed a method of CNN-based RSRL for the photos' aesthetics imbalanced classification. The networks are retrained repetitively by the self-revised training data set. Self-revision is done by dropping out the low likelihood photo samples scored in the middle levels from the training data set based on the previously trained network. Experimental results verified that the proposed method is effective for solving the issues of auto-assessment of photo aesthetics in imbalanced classification, and could be expected to extract the photos' highlight regions related with the aesthetic assessment by the repetitively retrained networks.

Moreover, we think that the proposed method is also available for other domains which are relevant to the imbalanced classification.